\documentclass[letterpaper, 10 pt, conference]{ieeeconf}  %

\IEEEoverridecommandlockouts       

\usepackage[backend=biber,
            url=false,
            isbn=false,
            doi=false,
            backref=false,
            style=ieee,
            natbib=true,%
            mincitenames=1,
            maxcitenames=1,
            citestyle=numeric-comp,
            sorting=nyt,%
            block=none]{biblatex}

\renewcommand{\bibfont}{\small}

\usepackage{algorithm}
\usepackage[noend]{algpseudocode}

\usepackage{graphics}
\usepackage{dsfont}
\usepackage[pdftex]{graphicx}
\usepackage{wrapfig}
\DeclareGraphicsExtensions{.pdf,.png,.jpg}
\usepackage{epsfig}
\usepackage[font={small}]{caption}
\usepackage{subcaption}
\usepackage[rightcaption]{sidecap}
\usepackage{pbox}

\usepackage{bigstrut}
\usepackage{mathtools}
\usepackage{amsmath, amssymb, amscd}
\usepackage{ wasysym } %
\usepackage{mathptmx} %
\usepackage{gensymb} 
\usepackage{nicefrac}       %
\numberwithin{equation}{section} 

\DeclareMathAlphabet{\mathcal}{OMS}{lmsy}{m}{n}
\usepackage{textcomp} %

\usepackage{algorithm} %
\usepackage{algorithmicx, algpseudocode} %

\usepackage{array} %
\usepackage{tabularx}
\usepackage{multirow}
\usepackage{multicol}
\usepackage{booktabs}
\usepackage{tabulary}

\usepackage[utf8]{inputenc}
\usepackage[english]{babel} %
\usepackage{units}
\usepackage{bm}
\usepackage{xspace}
\usepackage{flushend}%
\usepackage{balance} 
\usepackage{csquotes}
\usepackage{makeidx}
\usepackage{blindtext}

\let\labelindent\relax
\usepackage{enumitem}

\usepackage{ragged2e}
\usepackage{soul} %
\usepackage{subfiles} %

\usepackage{url}
\makeatletter
\g@addto@macro{\UrlBreaks}{\UrlOrds}
\makeatother
\usepackage{color}
\usepackage[usenames,dvipsnames,table,xcdraw]{xcolor}
\usepackage{pgfplots} %
\pgfplotsset{compat=newest}

\usepackage{marginnote}
\usepackage{soul} %

\usepackage[colorinlistoftodos]{todonotes}
\newcommand{\tocite}[1]{%
\textcolor{red}{[cite:\ifthenelse{\equal{#1}{}}{}{#1}?]}
}

\newcommand{\ignore}[1]{}

\renewcommand{\baselinestretch}{1.0}



\usepackage[pdfborder={0 0 0.5}]{hyperref}
\hypersetup{
    colorlinks=true,
    linkcolor=black,
    citecolor=black,
    filecolor=cyan,
    urlcolor=black
}

\title{\LARGE \bf
Learning Rope Manipulation Policies Using Dense Object Descriptors Trained on Synthetic Depth Data
}

\author{Priya Sundaresan$^{1}$, Jennifer Grannen$^{1}$, Brijen Thananjeyan$^{1}$, Ashwin Balakrishna$^{1}$, \\ Michael Laskey$^{2}$, Kevin Stone$^{2}$, Joseph E. Gonzalez$^{1}$, Ken Goldberg$^{1}$ 
\thanks{$^{1}$AUTOLAB at the University of California, Berkeley}
\thanks{$^{2}$Toyota Research Institute}%
}

\begin{document}

\maketitle

\begin{abstract}
Robotic manipulation of deformable 1D objects such as ropes, cables, and hoses is challenging due to the lack of high-fidelity analytic models and large configuration spaces. Furthermore, learning end-to-end manipulation policies directly from images and physical interaction requires significant time on a robot and can fail to generalize across tasks. We address these challenges using interpretable deep visual representations for rope, extending recent work on dense object descriptors for robot manipulation. This facilitates the design of interpretable and transferable geometric policies built on top of the learned representations, decoupling visual reasoning and control. We present an approach that learns point-pair correspondences between initial and goal rope configurations, which implicitly encodes geometric structure, entirely in simulation from synthetic depth images. We demonstrate that the learned representation — dense depth object descriptors (DDODs) — can be used to manipulate a real rope into a variety of different arrangements either by learning from demonstrations or using interpretable geometric policies. In 50 trials of a knot-tying task with the ABB YuMi Robot, the system achieves a 66\% knot-tying success rate from previously unseen configurations. See \url{https://tinyurl.com/rope-learning} for supplementary material and videos.

\end{abstract}

\section{Introduction}
\label{sec:introduction}
Manipulating deformable objects is valuable for a wide variety of applications from surgery and manufacturing to household robotics~\cite{superhuman-surgery, SAVED, robotic-heart-surgery, dense-obj-nets, grasp2vec, bed-making, thananjeyan2017multilateral, grasping-deformable-obj, feeling-the-grip, interactive-comp-imaging}. We specifically consider manipulation of rope, whose infinite dimensional configuration space objects makes it difficult to build accurate dynamical models.  Rope manipulation is also difficult because of significant perception challenges due to self occlusions, loops, and self-similarity \cite{knot-theory}. There has been prior work successfully utilizing finite element models~\cite{flexible-objects} and hard-coded representations for deformable manipulation~\cite{learning-by-watching, knotting, knot-planning, towel-folding}, but these techniques can fail to generalize to novel configurations.

\begin{figure}[t!]
  \includegraphics[width=0.45\textwidth]{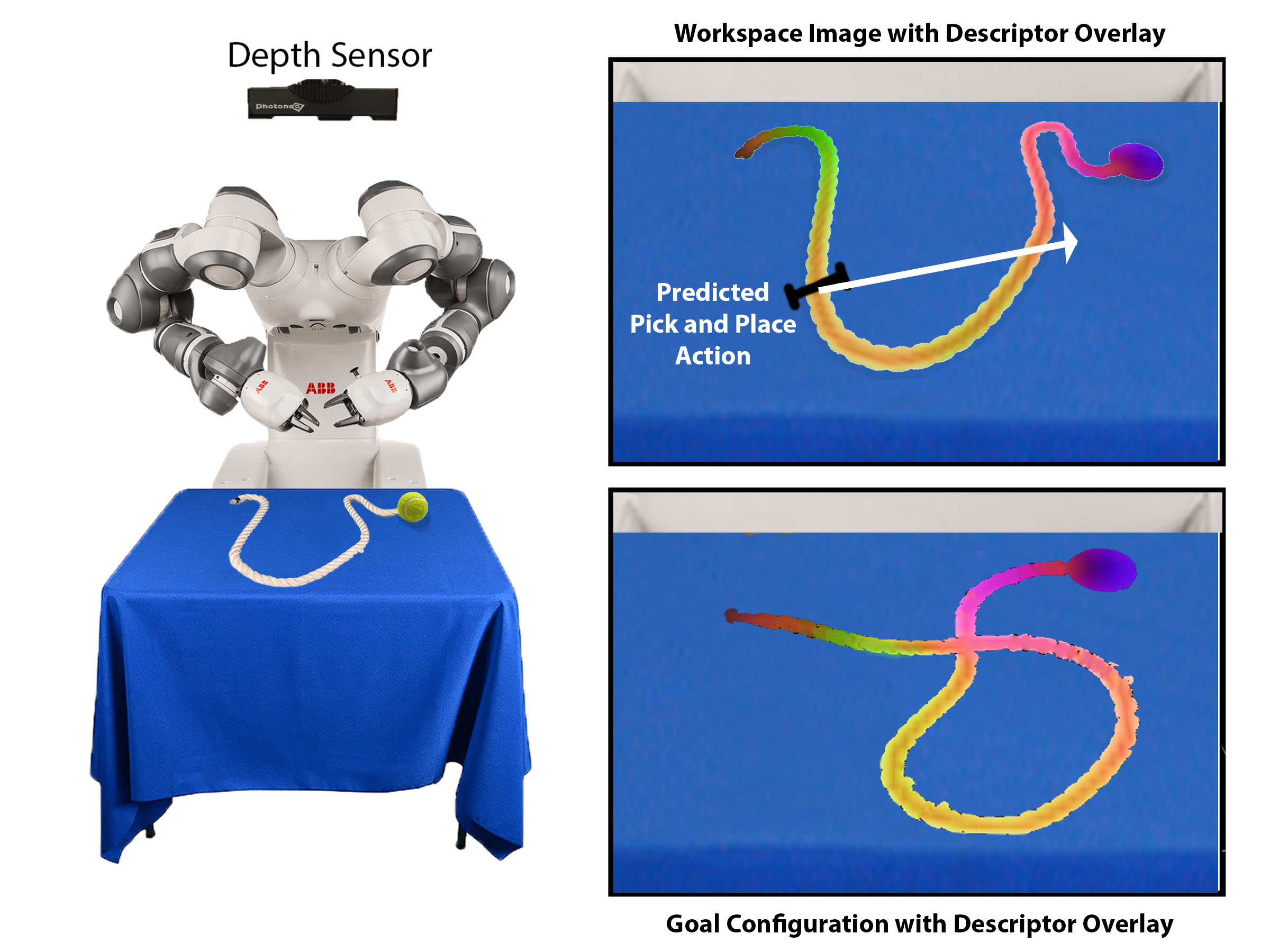}
  \caption{The robot uses dense depth object descriptors (DDODs), learned from synthetic depth images, to compare its current depth observation to a depth image of the desired configuration and plan actions to guide the rope to the goal configuration. We use this strategy to track video demonstrations of rope manipulation tasks and to define a geometric algorithm that ties knots from previously unseen starting configurations. A ball is added to the rope to break symmetry and enable consistent correspondence mapping. Although we exclusively use depth images for training and recording observations during manipulation, we show color images of the workspace for visual clarity. } 
  \label{splash-fig}
\end{figure}

\begin{figure}[t!]
\centering
  \includegraphics[width=0.44\textwidth]{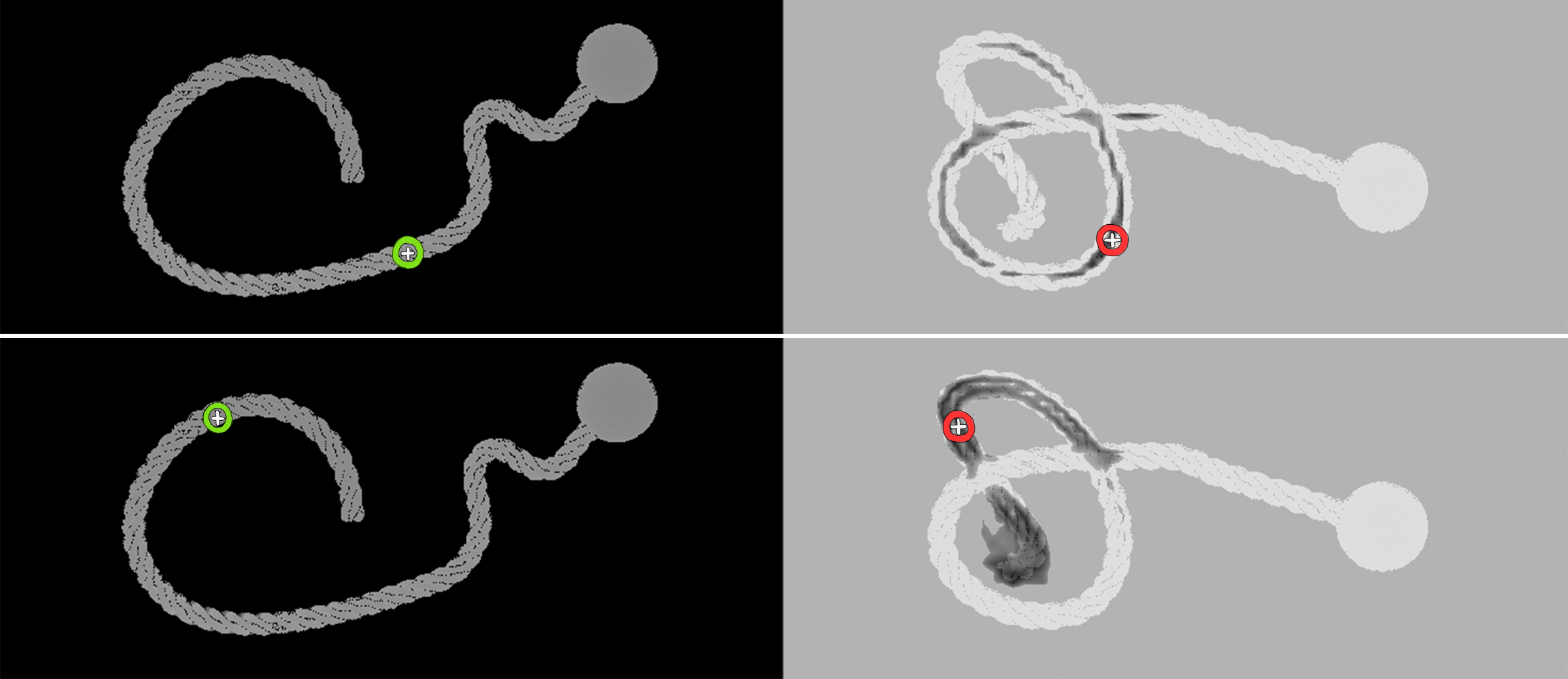}
  \caption{A visualization of learned descriptors, where the right column images display predicted pixel correspondences (red cursors) relative to the left image source pixels (green cursors) and predicted best match regions (darkened) \cite{dense-obj-nets}. This is generated by applying the learned descriptor mapping: $\psi: \mathbb{R}^{W \times H \times 1}_+ \rightarrow \mathbb{R}^{W \times H \times K}_+$ independently to both synthetic depth images, computing the pixelwise norm differences in descriptor space, and scaling these differences linearly $\in$ [0, 255]. The darkened regions can be interpreted as a measure of uncertainty in predicted correspondences. Note that the predicted correspondences are sensitive to self-intersections.} 
  \label{heatmap-fig}
\end{figure}

These perception and modeling challenges motivate learning-based strategies. Past learning-based approaches have achieved impressive results on a variety of rope manipulation tasks, but require many hours of real-world data collection to learn action-conditioned visual dynamics models of the rope~\cite{vision-based-rope-manip, visual-planning-acting, zero-shot-visual-IL}. We address these issues by decoupling perception from planning and control. We learn abstract visual representations of rope by extending the techniques from \cite{visual-descriptors,dense-obj-nets} to learn descriptors for the rope that are invariant across different configurations (Figure \ref{heatmap-fig}). We then demonstrate that these representations can be leveraged to create both \emph{interpretable} (visually intuitive and geometrically structured) and \emph{transferable} polices (task agnostic, learned from synthetic images, deployed on real images) for achieving various planar and non-planar rope configurations (Figure \ref{splash-fig}). Shifting the representational load from the control policy to a separate perception module enables learning to encode information about rope geometry in simulation without real data. Furthermore, because the object descriptors are trained only on images of the rope in different configurations and are agnostic to the actions that generated them, accurate dynamic simulation of the rope is unnecessary.
This paper provides four contributions: (1) a novel approach to achieve complex planar and non-planar rope configurations with a single video demonstration of the task by tracking the learned \emph{dense depth object descriptors} (DDODs); (2) experiments suggesting that the dense object descriptors from~\citet{visual-descriptors, dense-obj-nets}, previously applied to learn representations for rigid bodies and slightly deformable objects using real data, can be extended to learning representations for highly deformable objects such as rope using only synthetic depth images; (3) a geometrically-motivated algorithm using DDODs to tie knots from unseen rope configurations; and (4) experiments with an ABB YuMi robot suggesting the learned DDODs can be used to achieve a set of planar/non-planar rope configurations and successful knot-tying in 33/50 trials from previously unseen states.

\section{Background and Related Work}
\label{sec:related-work}

There is recent work on tracking deformable objects in videos such as \cite{non-rigid-registration, hand-tracking, visual-descriptors, dynamic-fusion, online-deformable-tracking, interactive-comp-imaging}. There is also extensive literature on deformable manipulation~\cite{learning-by-watching, knotting, knot-planning, towel-folding, LfD-non-rigid}. We primarily focus on learning-based methods, which have been shown to generalize to a variety of tasks~\cite{vision-based-rope-manip, visual-planning-acting, zero-shot-visual-IL}. Due to the challenge of designing accurate analytical models for deformable objects, \cite{vision-based-rope-manip, visual-planning-acting, zero-shot-visual-IL} provide effective learning-based algorithms for rope manipulation by either generating a visual plan or using an existing one from demonstrations, and then executing the plan by generating controls using learned dynamics models given a single video demonstration. However, these methods require tens of hours of real data collection to learn rope dynamics. These approaches also do not impose any geometric structure on the learned visual representations, limiting the interpretability of the learned policies. In contrast, we impose geometric structure on the learned visual representations via DDODs, learn them in simulation, and decouple them from robot actions. This accelerates training time substantially, and makes it easier to transfer the learned visual representation across domains.

We learn geometrically meaningful visual representations for rope by using dense object descriptors, introduced in the context of robotic manipulation by \cite{dense-obj-nets}. While task agnostic manipulation requires geometric understanding of the objects being manipulated, fine-grained understanding of the object configuration is often unnecessary to effectively grasp or push an object~\cite{dexnet1, dexnet2, dexnet3, linear-pushing, grasping-deformable-obj, feeling-the-grip}. We leverage dense descriptors for task-oriented manipulation, which often requires detailed geometric understanding to manipulate objects in the specific ways needed to achieve task success \cite{dense-obj-nets, task-oriented-semantic}. There exists extensive literature on generating descriptors for keypoints in images \cite{SIFT, HOG}, but these approaches rely on image intensity gradients, which will not provide much signal in images where the pixel intensities and textures are largely homogeneous such as for a rope. This motivates a deep learning-based approach to utilize global information about the rope to generate descriptors and correspondences~\cite{visual-descriptors, dense-obj-nets, compact-geometric, pose-reg}.

~\citet{visual-descriptors} propose a deep learning approach to learn a function that maps pixels corresponding to the same point on an object to the same descriptor and pixels corresponding to different points to different descriptors.~\citet{dense-obj-nets} use these dense object descriptors for task-oriented manipulation of rigid and slightly deformable objects such as stuffed animals. In contrast to prior work, we demonstrate that similar descriptors can be learned and leveraged for manipulation of very deformable 1D structures such as rope. We also learn descriptors from While \cite{visual-descriptors, dense-obj-nets, compact-geometric} learn descriptors using color image input, we use synthetic depth input, which facilitates sim-to-real transfer of the learned representations ~\cite{bed-making, dexnet2} and richly encodes the geometric structure of ropes in knotted configurations.

\section{Simulator}
\begin{figure*}[t!]
\centering
  \includegraphics[width=0.9\textwidth]{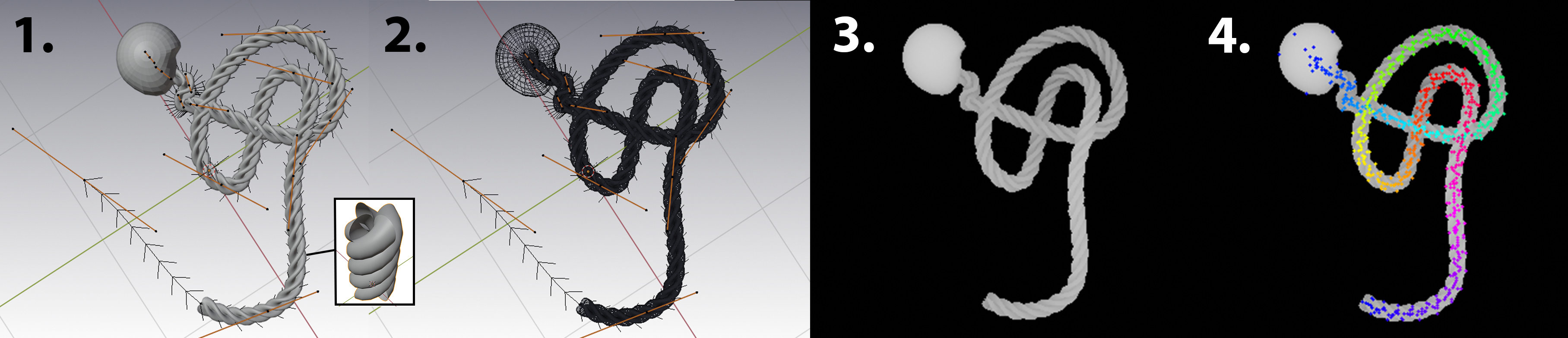}
  \caption{Rope simulation design. 1) The underlying representation of the rope is a set of $M$=12 Bezier control points (visualized as black points with orange handles). These nodes can be randomly displaced along x, y, or z axes to produce arbitrary deformation or can be fixed according to a control polygon to produce structured deformation such as loops, overlaps, and knots. The Bezier curve is of variable length while the rope mesh is of fixed length. The Bezier nodes may become unequally spaced during displacement, as shown in sub-figure 1 when only 7 of the 12 nodes are visible after deformation. 2) The wireframe rope mesh with ordered vertices of known coordinates. 3) A rendered depth map. 4) A visualization of the densely annotated scene with $N$=1,465 pixels corresponding to $N$ vertices sampled from the rope mesh in 3). The pixels are colored in a stream to demonstrate the ordering of the dense ground truth annotations in simulation.  } 
  \label{simulator-fig}
\end{figure*}

\label{sec:simulator}
We use Blender 2.8 \cite{roosendaal2007blender} — an open-source 3D graphics, animation, and rendering suite — to model the rope in simulation and generate synthetic depth training data. Hyperparameter details for the simulation environment are provided in Section \ref{sec:appendix} (Table \ref{tab:blender-params}). The simulated rope is modelled by twisting four thin cylindrical meshes to produce a realistic braided twine appearance as in \cite{blendertutorial}. A sphere mesh was added on one end to break the symmetry of the rope, which was experimentally shown to reduce ambiguity in descriptor learning. This rope representation consists of a mesh with over fifty thousand ordered vertices of known global coordinates and an underlying Bezier curve with $M = 12$ control points, ${\textbf{P}_1, ..., \textbf{P}_\textnormal{M}}$ (Figure \ref{simulator-fig}). A larger $M$ value enables higher manipulation fidelity and a larger configuration space for the rope. Simple configurations consist of purely planar deformations, formed by picking random points along the rope and pulling arbitrarily along the $x$ and $y$ directions. Complex configurations include planar deformations in addition to randomized overlap, loops, and knots.  Producing varied synthetic depth training data requires simulating the rope in a variety of configurations and exporting the relevant ground truth data and rendered image. For the first step, we randomize the positions of a subset of the Bezier control points to produce varied deformations. Next, for a given scene, we export a depth image from the scene's Z-Buffer output and a mapping $i \rightarrow (u_i, v_i), i \in (1, ..., N)$. This represents the projection of $N$ vertex world coordinates to pixel coordinates in the synthetic camera frame. The parameter $N$ specifies how many pixels to annotate on the image, so a higher value of $N$ produces more dense pixel match sampling between images during training. This raw projection mapping fails to account for complex rope geometries, since multiple mesh vertices can project to the same pixel coordinate at regions of self-intersection or occlusion. Thus, we reparent all pixels in a given region to the top-most mesh vertex in that region using a $k$-nearest neighbor algorithm with $k=4$. That is, given $(u_j, v_j)$ and $(u_k, v_k)$ such that $||(u_j, v_j) - (u_k, v_k)||_2$ $\leq \gamma$, we compare the z-coordinates of the corresponding mesh vertex world coordinates, $p_j = (x_j, y_j, z_j)$ and $p_k = (x_k, y_k, z_k)$, respectively. If $z_k > z_j$, the exported mapping will assign both $[(u_k, v_k), (u_j, v_j)]$ to $k$ instead of $j$. Pixel matches can be sampled across images of varying configurations by pairing pixels by corresponding mesh vertex. 

\section{Dense Descriptor Learning}
Here we describe the training procedure for training dense object descriptors for rope manipulation from synthetic depth data. Hyperparameters for descriptor learning are specified in Appendix~\ref{sec:appendix} (Table \ref{tab:training-params}).
\label{sec:descriptor-learning}

\subsection{Preliminaries}
\label{sec:preliminaries}
\begin{figure}[t!]
\centering
  \includegraphics[width=1.0\linewidth]{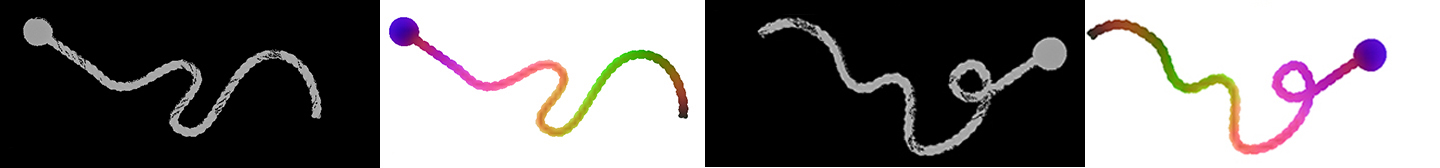}
  \caption{A visualization of trained, normalized rope descriptors applied to synthetic depth images unseen during training. The first and third images show examples of synthetic depth images of a rope in different configurations. The second and fourth represent the output of the dense correspondence network, where for each pixel on the rope mask, the normalized 3D descriptor vector is visualized as a RGB tuple. The visualizations suggest descriptor consistency across deformations.} 
  \label{descriptor-vis}
\end{figure}

We consider an environment which consists of a static flat plane and a braided rope and learn policies to achieve specific planar and non-planar configurations. We do this by learning a structured visual representation of the rope to estimate point-pair correspondences between an overhead depth image of the rope and a subgoal image. These correspondences are then used to generate interpretable geometric policies which move the rope to better align it with the subgoal. For more details on how the policies are defined, see Section \ref{sec:policy-generation}.

For visual representation learning, we build on the work in \cite{visual-descriptors, dense-obj-nets} by learning descriptors from depth images in addition to RGB and extending the framework to a highly deformable object. In \citet{dense-obj-nets}, representation learning is done by first sampling a variety of points on the surface a given object. The camera pose is changed via a randomly sampled rigid body transformation and the sampled points are associated with corresponding points in the new view using standard static scene reconstruction techniques. These correspondences are then used to train a Siamese network~\cite{siamese} with pixelwise contrastive loss to learn the desired embedding space. See \cite{dense-obj-nets} for more details. \citet{dense-obj-nets} demonstrate that these descriptors can be used to pick up rigid and slightly deformable objects at specific grasp points from multiple views, even when the target grasp is only identified in one view. Unlike \cite{dense-obj-nets}, since the rope is not rigid, it is insufficient to simply change the pose of the camera to learn object descriptors for manipulation. Thus, the rope must be manipulated into a variety of different possible configurations to generate useful correspondences. Since ground truth correspondences are difficult to obtain for a real rope, we leverage simulation to obtain point-pair correspondences, which are then used to learn DDODs. Unlike \cite{dense-obj-nets}, which train descriptors on RGB images, we train on synthetic depth~\cite{dense-obj-nets}.

\subsection{Descriptor Learning from Synthetic Depth Images}
\label{sec:descriptor-learning-synthetic-data}
The training procedure involves sampling a random initial configuration of the rope $\xi_1$ in simulation and applying some transformation $\phi$ to yield a new configuration $\xi_2$. As in \citet{dense-obj-nets}, the goal is to learn a mapping to a descriptor space in which corresponding points on $\xi_1$ and $\xi_2$ are encouraged to be close together while non-corresponding points are encouraged to be further apart. 

\begin{figure}[t!]
\centering
  \includegraphics[width=0.4\textwidth]{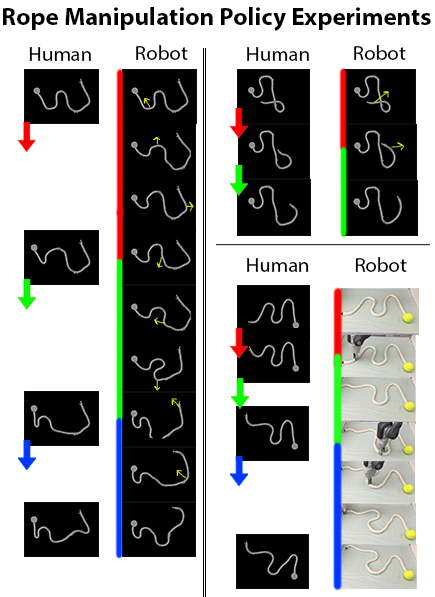}
  \caption{Three examples of rope manipulation action sequences the YuMi robot performed by one-shot visual imitation of a demonstrated sequence of observations. Each demonstrated sequence consists of a starting configuration followed by pick-and-place actions performed by a human supervisor to produce a different final state. For each step in the demonstration, the YuMi is given a fixed number of pick-and-place attempts (1 for non-planar sequences, 3 for planar sequences) to produce the next sequential state, unless the IoU of the current workspace image and the goal state is below a hand-tuned threshold (0.67). We allow fewer attempts for the non-planar case because we observed that it is more difficult for the robot to recover from poor nonplanar actions since these often produce entanglement or particularly pathological configurations, whereas missteps in the planar actions sequence are typically less costly since the rope is likely to remain planar and correspondences can be resampled. For a single action, the YuMi executes a greedy policy by grasping the correspondence on the rope in the current image that is farthest from its pixelwise match in the goal image and placing it at that point. Qualitative results suggest the efficacy of the geometric policy defined over the learned descriptors.} 
  \label{planar-exp-fig}
\end{figure}

We generate planar transforms by randomly translating the coordinates of a subsample of the rope's Bezier knots ${\textbf{P}_1, ..., \textbf{P}_\textnormal{M}}$ along the x and y axes to simulate pulling the rope arbitrarily along different directions. We also generate transforms that simulate more complex rope configurations including overlap, loops, and knots by geometrically arranging ${\textbf{P}_1, ..., \textbf{P}_\textnormal{M}}$ into the respective control polygons for these configurations as in \cite{ma2003point}, and then slightly perturbing knot coordinate positions for variation. 

We sample a set of $N$ corresponding point pairs $p = (p_{1i}, p_{2i})_{i=1}^{N}$ on the rope between configurations $\xi_1$ and $\xi_2$.  This allows us to sample a wide variety of possible rope deformations, making it easier to generalize to different tasks at test-time. Learning in simulation also makes it possible to inject noise to enable robustness to varying experimental conditions as described in Section \ref{sec:sim2real}. Then, we utilize the same training procedure as in \cite{dense-obj-nets} to learn $K$-dimensional DDODs, where $K$ is a hyperparameter that we experimentally vary between 3 and 16.

\section{Policy Design}
\label{sec:policy-learning}
\label{sec:policy-generation}
Given the learned descriptors, we design interpretable geometric policies defined over the learned DDODs. We assume that the rope manipulation tasks considered can be performed by a sequence of pick and place actions by a single robotic arm as in prior work~\cite{vision-based-rope-manip, zero-shot-visual-IL}. Hyperparameter details regarding manipulation policies are specified in Appendix~\ref{sec:appendix} (Table \ref{tab:manip-params}). We consider two algorithmic policies for rope manipulation tasks:
\subsection{Algorithm 1: One-Shot Visual Imitation}
\label{sec:imitation-policy}
In this setting, a human demonstrator makes sequential pick and place actions to arrange the physical rope into a desired configuration. The robot observes one demonstration as a sequence of images from an overhead depth camera, then takes actions based on a greedy geometric policy.

Actions, defined by a start point (grasp) $p_s \in \mathbb{R}^2$ and an end point (drop) $p_e \in \mathbb{R}^2$, are generated by using the frames in the provided demonstration as subgoals and using the descriptors to sparsely estimate point-pair correspondences between points on the current depth image of the rope at time $t$ and the current subgoal, given by a demonstration frame (Figure \ref{planar-exp-fig}). To find correspondences, we sparsely sample a set of roughly evenly spaced pixels on the rope mask in the current depth image by enforcing the constraint that the inter-pixel distance between any two points should be above a margin $\alpha = 50$. For each of the sparsely sampled pixels, we compute their correspondence on the goal image by computing the $100$ nearest neighbors in descriptor space and taking the best match to be the median of the associated $100$ pixels. We choose the median correspondence due to its robustness to outliers.

Then, we find the pair of corresponding points with the highest discrepancy (largest distance in $\mathbb{R}^2$ between them), and take the following action to align these points in 3D space: the point-pair correspondence $(p_{1}, p_{2})$ with the maximum discrepancy is computed and the robot grasps the rope at point $p_{1}$ and places the rope at point $p_{2}$ to align the furthest points in the image. This process is repeated up to $k$ times for each subgoal image or until the intersection-over-union (IoU) of the current and goal state image masks is below a hand-tuned threshold of 0.67. The IoU is a standardized metric across segmentation tasks~\cite{mask-rcnn} and provides an indication of the degree of alignment between two masks, which we use to judge the similarity of two rope configurations. We found the IoU to be a noisy measurement for alignment of current and subgoal rope masks, and use a relatively low threshold to account for this. This is likely caused by the long, thin geometry of the rope, which complicates pixelwise alignment of two otherwise very similar rope configurations. 

\begin{figure}[t!]
  \includegraphics[width=1.0\linewidth]{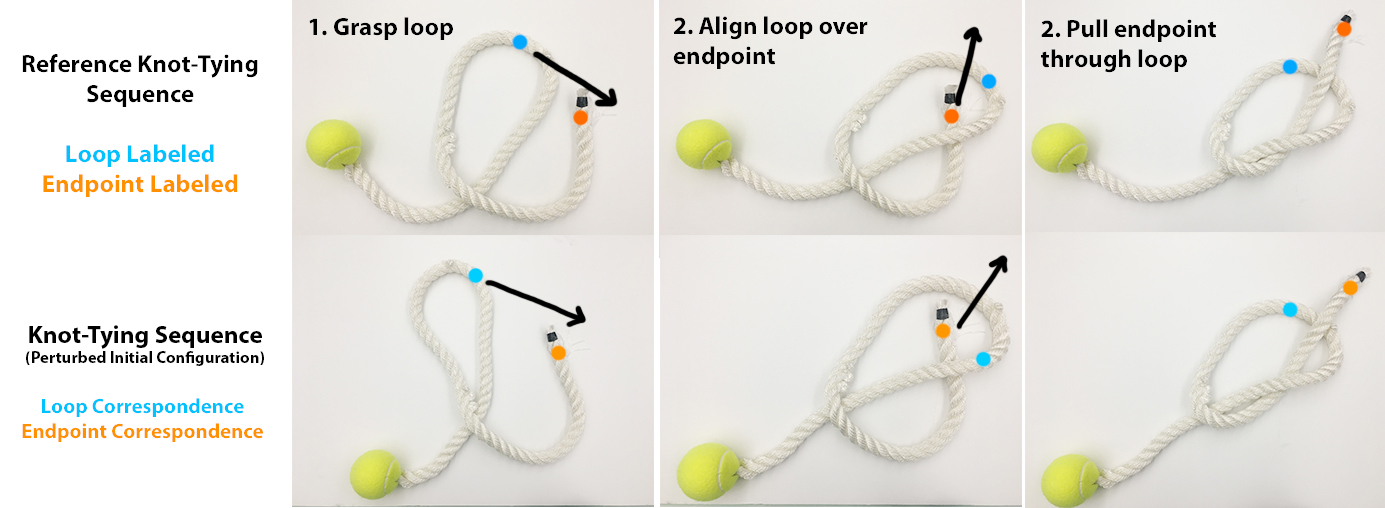}
  \caption{To perform knot-tying, we label the centered loop point and endpoint of the rope in a reference image, and define two geometric pick-and-place actions in terms of the relative spacing of these points to generate a knot. To generalize to a new initial loop configuration, we recompute loop and endpoint correspondences and execute the sequence.} 
  \label{knot-fig}
\end{figure}

\subsection{Algorithm 2: Descriptor Parameterized Knot-Tying}
\label{sec:descriptor-alg}
In this setting, we use a two-action sequence of a knot-tying task from a human demonstrator to parameterize a sequence of motion primitives for knot-tying that generalizes to unseen rope configurations. As in \cite{vision-based-rope-manip}, we assume the rope contains a single loop initially. The sequence is annotated with the two pick and place actions used to execute the task (Figure~\ref{knot-fig}).

The first action involves picking the side of the loop close to the end of the rope without the ball and placing it around the endpoint of the rope. We record the descriptor vectors for the grasp point and the end of the rope and use it to define an action in terms of DDODs. When faced with a new, unseen rope configuration with a loop, the robot grasps the closest point in descriptor space to the grasp point in the reference image and pulls it in the direction of the end of the rope, which is also found by matching with the closest descriptor in the reference frame.

The next step involves grasping the end of the rope in the loop and pulling it to tighten the knot. To define this primitive, we record the descriptor vector for the end of the rope in the reference image. When executing this maneuver in a new configuration, the robot detects the end of the rope by finding the closest pixel in descriptor space to the end of the rope in the reference image. The robot grasps at this point and pulls to tighten the knot.
\begin{figure*}[t!]
\centering
  \includegraphics[width=0.99\textwidth]{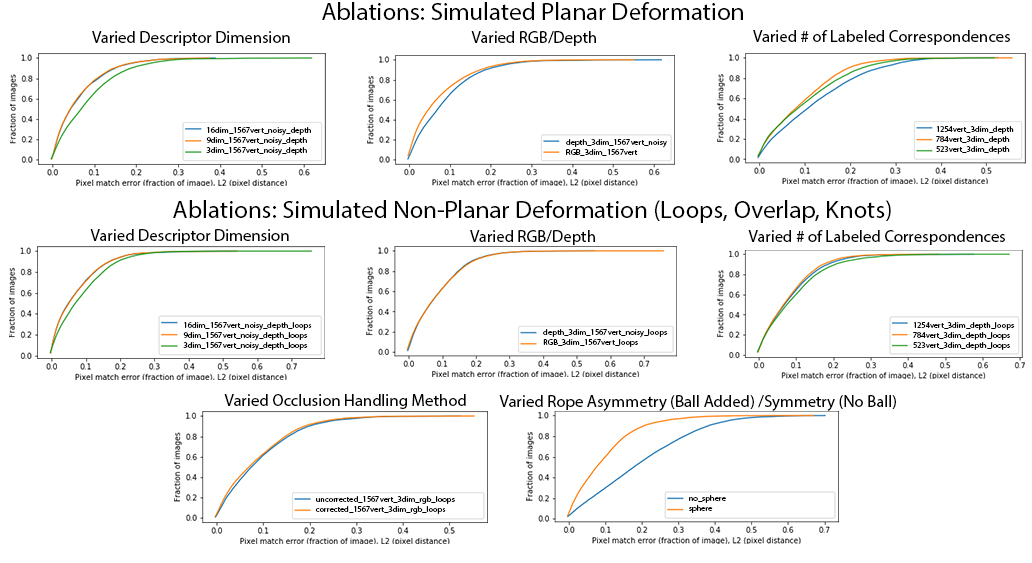}
  \caption{Ablations measuring pixel-match error for the learned descriptors in simulation when descriptor dimension, sensing modality, number of correspondences used for training, and occlusion handling method are varied. Results suggest that the learned representations are largely insensitive to small changes in these parameters, with the exception of adding a ball to the end of the rope. Asymmetry is critical for good performance, as removing the ball results in a significant deterioration in performance as expected. Furthermore, we note that depth input performs nearly identical to RGB for non-planar configurations, which is consistent with the increased depth variation in non-planar settings.}
  \label{ablations-fig}
\end{figure*}

\section{Experiments}
\label{sec:results}
\vspace{-0.25cm}
\subsection{Baseline}
We propose an analytical method for acquiring rope correspondences and performing manipulation which we compare against the dense correspondence method. This analytical method is detailed in the Appendix (Section \ref{sec:appendix}) and relies on following the pixel intensity gradient of the rope at local crops to collect annotations along the rope geometry. This method is largely hand-tuned to the rope images observed by the depth camera and lacks robustness to small irregularities in the rope such as fraying and non-uniform thickness. It is also currently unsupported for the non-planar case, as following the gradient of the rope is nontrivial when the rope overlaps on itself. These challenges with designing an analytical baseline motivate the DDOD-based approach. 
\subsection{Experimental Setup}
We use a 4 ft. by 1/2 in. braided white nylon rope with a punctured tennis ball attached to one end to resolve ambiguity between the ends of the rope and to match the appearance of the rope in simulation. 
We assume access to observations from a overhead depth camera (Photoneo Phoxi 3D Scanner) and visibility of the rope in its entirety (including endpoints) throughout the duration of the task. We further assume a relatively flat background with no distractor objects. In this experimental setup, it is infeasible for the robot to do one shot visual imitation of the human demonstrator in sequence, since the rope state is changed from its initial configuration by the end of the demonstration. Thus, we also assume that the robot can start from the last recorded demonstration frame and do one shot visual imitation in reverse to restore the original configuration of the rope. Additional details about the experimental setup are provided in the Appendix (Section \ref{sec:appendix}).

\subsection{Simulated Experiments}
In simulation, we train the deep network used in \citet{dense-obj-nets} to learn point-pair correspondences for a variety of rope deformations as described in Section \ref{sec:descriptor-learning}, for both simple and complex tier configurations. For each network, we train on a set of 3,600 generated synthetic depth and RGB images (~1 hr. data generation time) and evaluate on a held-out test set of 100 pairs of previously unseen images. Descriptor quality is measured in terms of pixel-match error on the held-out test set as in \cite{dense-obj-nets}. Experiments suggest that the learned descriptors are consistent and able to accurately locate correspondences in images of rope in unseen configurations. Figure \ref{descriptor-vis} shows a few qualitative examples. 

In Figure~\ref{ablations-fig}, we evaluate the quality of the learned descriptors when we vary the sensing modality (synthetic RGB/synthetic depth), the descriptor dimension, the number of annotated correspondences, and when we ignore/account for occlusions in the nonplanar datasets using the method described in Section \ref{sec:simulator}. We see that the descriptor quality is largely invariant to small changes in descriptor dimension, sensing modality, annotation density, and occlusion handling. For non-planar deformations, the gap in the pixelwise error for descriptors trained on RGB and depth data is observed to be significantly lower than for planar deformations. Given the greater depth variation in images, depth data is likely more rich and useful in the nonplanar case. We also observe the benefit of the added ball for breaking symmetry.

 \begin{table}[!htbp]

\centering
\resizebox{\columnwidth}{!}{
 \begin{tabular}{||c || c || r || r ||} 
 \hline
 Type & Subgoal & Trials w/ Improvement & Med. \% Improvement  \\ 
 \hline\hline
Baseline (P) & 0 & 9/9 & 56\\
 \hline
Baseline (P) & 0 & 4/9 & -4\\
 \hline
Baseline (P) & 0 & 6/9 & 42\\
 \hline
DDOD (P) & 0 & 28/32 & 58\\ 
 \hline
DDOD (P) & 1 & 28/32 & 42\\ 
 \hline
DDOD (P) & 2 & 23/32 & 33\\ 
 \hline
 
DDOD (NP) & 0 & 14/21 & 30\\ 
 \hline
DDOD (NP) & 1 & 13/21 & 4\\ 
\hline
\end{tabular}}
\caption{Physical Experiment Results (Visual Imitation): We report the number of trials that improve with respect to the subgoal-based loss defined in Section~\ref{sec:visual-imitation-exps} for planar (P) and non-planar (NP) visual imitation experiments. We find that even in the non-planar case, the robot makes positive progress in most trials, but note that performance decreases as the task progresses. We also report the median percent improvement of the loss over each subgoal's starting configuration. We report the median, because failures cause large negative outlier loss values, skewing the mean. We find that the visual imitation policy using dense object descriptors is able to drive the rope to configurations closer to the target configurations. We observe that performance deteriorates in later subgoals, which we hypothesize is due to compounding errors over time. We observe that non-planar manipulation is more challenging.}
\label{table:imitation}
 \end{table}

\vspace{-0.5cm}
\begin{table}[!htbp]
    \caption{Classification of the 33\% Failures for Physical Knot Tying}
    \centering
    \begin{tabular}{||c || l || r||} 

 \hline
 Mode & Explanation & Count  \\ 
 \hline\hline
A & wrong endpoint correspondence & 4 \\ 
 \hline
B & wrong loop point correspondence & 6 \\ 
 \hline
C & endpoint occluded after pull & 3 \\ 
 \hline
D & loop pulled misaligned & 3 \\ 
 \hline
\end{tabular}
\label{tab:my_label}
\end{table}
\vspace{-0.5cm}

\subsection{Physical Experiments}
We evaluate the learned representations for designing rope manipulation policies with an ABB YuMi robot equipped with one parallel jaw gripper. For planar physical experiments, we use a 3-dimensional DDOD network trained on simulated planar configurations with 1,400 labeled correspondences per rendering. For nonplanar manipulation, we use a 16-dimensional DDOD network trained on simulated nonplanar configurations with 557 annotations per rendering. Both networks are trained on noise-injected simulation images (Appendix, Section \ref{sec:sim2real}) to enable transfer to the real rope. We use the networks to perform manipulation using the geometric policies from Section \ref{sec:policy-generation}.

\subsubsection{Alg 1}
\label{sec:visual-imitation-exps}
We evaluate Algorithm 1 on its ability to track and repeat video sequences of both planar and non-planar rope manipulation as shown in Figure \ref{planar-exp-fig} and Table \ref{table:imitation}. Each planar and non-planar sequence consists of three or four frames respectively, including a starting configuration. For each of the subgoals, the robot executes up to $3$ or $1$ actions for planar and non-planar experiments respectively, and proceeds early to the next subgoal if the IoU threshold in Section~\ref{sec:imitation-policy} is met.

\paragraph{Evaluation Metric} To evaluate the agent's ability to track the subgoals in the video sequence, we define a loss function that takes in the realized image $I_{real}$ and the goal image $I_{goal}$: $L(I_{real}, I_{goal})$. For each image $I$, a sequence of points along the rope is manually annotated, and a parametric piecewise linear function $p_{I}(i)$ is fit to the points for $i\in[0, 1]$. Then, the sum of squared errors is computed for a range of shifts and rotations of $I_{real}$ for $100$ evenly spaced points on the curve and the minimum is returned by $L$. For each subgoal in the demonstration trajectory, $L$ is computed for all frames in the segment corresponding to it in the robot trajectory and report the percent improvement of the best frame over the segment's starting configuration (Figure~\ref{sec:visual-imitation-exps}, Table \ref{table:imitation}).

\subsubsection{Alg 2}
We evaluate the method in Section~\ref{sec:descriptor-alg} on a knot-tying task from $50$ previously unseen configurations with the rope starting in a loop. As in prior work~\cite{vision-based-rope-manip,zero-shot-visual-IL}, we report the success rate of the task by visually inspecting whether a knot was successfully tied. Figure \ref{knot-fig} illustrates the knot-tying procedure used. The robot successfully ties a knot in $33/50$ trials (66\%). This rate is higher than the knot-tying accuracy reported in \cite{vision-based-rope-manip} (38\%) and \cite{zero-shot-visual-IL} (60\%), and requires weaker supervision, although we do not provide a direct comparison due to differences in experimental setup. Failure modes include when the robot fails to accurately identify the loop and endpoint correspondences, fails to align the loop over the endpoint, or occludes the endpoint during alignment, preventing task completion (Table \ref{tab:my_label}).

\section{Discussion and Future Work}
\label{sec:discussion}
This work presents a new method for designing interpretable and transferable policies for rope manipulation by learning a geometrically structured visual representation (DDOD) entirely in simulation by building on the techniques from \citet{dense-obj-nets}. The visual correspondence-driven manipulation policies demonstrated allow for ease of interpretation and understanding of robotic actions in both a one-shot visual imitation framework and a descriptor-parameterized task setting. We use this representation to design intuitive geometric policies to track planar and non-planar rope deformations from demonstrations and to design a geometric algorithm for knot tying which achieves a 66\% success rate. In future work, we will explore learning more complex manipulation primitives in descriptor space such as suturing. We will also investigate whether the learned descriptors provide appropriate representations for reinforcement learning and for manipulation of 2D deformable objects like cloth.

\vspace{-0.5cm}
\section{Acknowledgments}
\footnotesize
This research was performed at the AUTOLAB at UC Berkeley with partial support from Toyota Research Institute, the Berkeley AI Research (BAIR) Lab
and by equipment grants from PhotoNeo. Any opinions, findings, and conclusions or recommendations expressed in this material are those of the author(s) and do not necessarily reflect the views of the sponsors. Ashwin Balakrishna is supported by an NSF GRFP. We thank our colleagues who provided helpful feedback, code, and suggestions, especially David Tseng, Aditya Ganapathi, Michael Danielczuk, Jeffrey Ichnowski, and Daniel Seita.

\printbibliography
\clearpage

\normalsize
\section{Appendix}
\label{sec:appendix}
The appendix is organized as follows:
\begin{itemize}
    \item Appendix A contains additional details on the rope simulator
    \item Appendix B describes the baseline policy for planar manipulation
    \item Appendix C contains additional details on the experimental setup
    \item Appendix D specifies hyperparameters for the simulator, descriptor training, and manipulation policies.
\end{itemize}
\subsection{Rope Simulator Details}
\begin{figure}[!htbp]
  \includegraphics[width=1.0\linewidth]{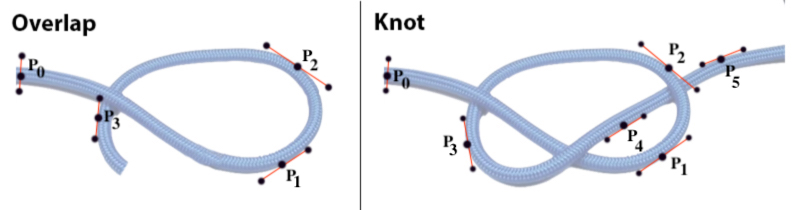}
\caption{Four Bezier control points are needed to parametrize an overlapping configuration, and six Bezier control points form the control polygon for a knotted configuration. In practice, we slightly randomize the positions of these control points to yield non-uniform nonplanar arrangements.  }
  \label{gt-fig}
\end{figure}

The rope simulator is implemented in the graphics/rendering engine Blender 2.8 using its Python API. We only use Blender's rendering capabilities, rather than its dynamic simulation capabilities, to produce varied training data of the rope in different configurations. Blender preserves the ordering of mesh vertices after deformation, so for each image rendered, we export dense, ordered pixel-wise annotations using the world-to-camera transform on queried mesh vertices (Figure \ref{gt-fig}). Finally, the rendered images are injected with noise to resemble real depth images.

\subsubsection{Rope Model} 
\begin{figure}[!htbp]
  \includegraphics[width=0.47\textwidth]{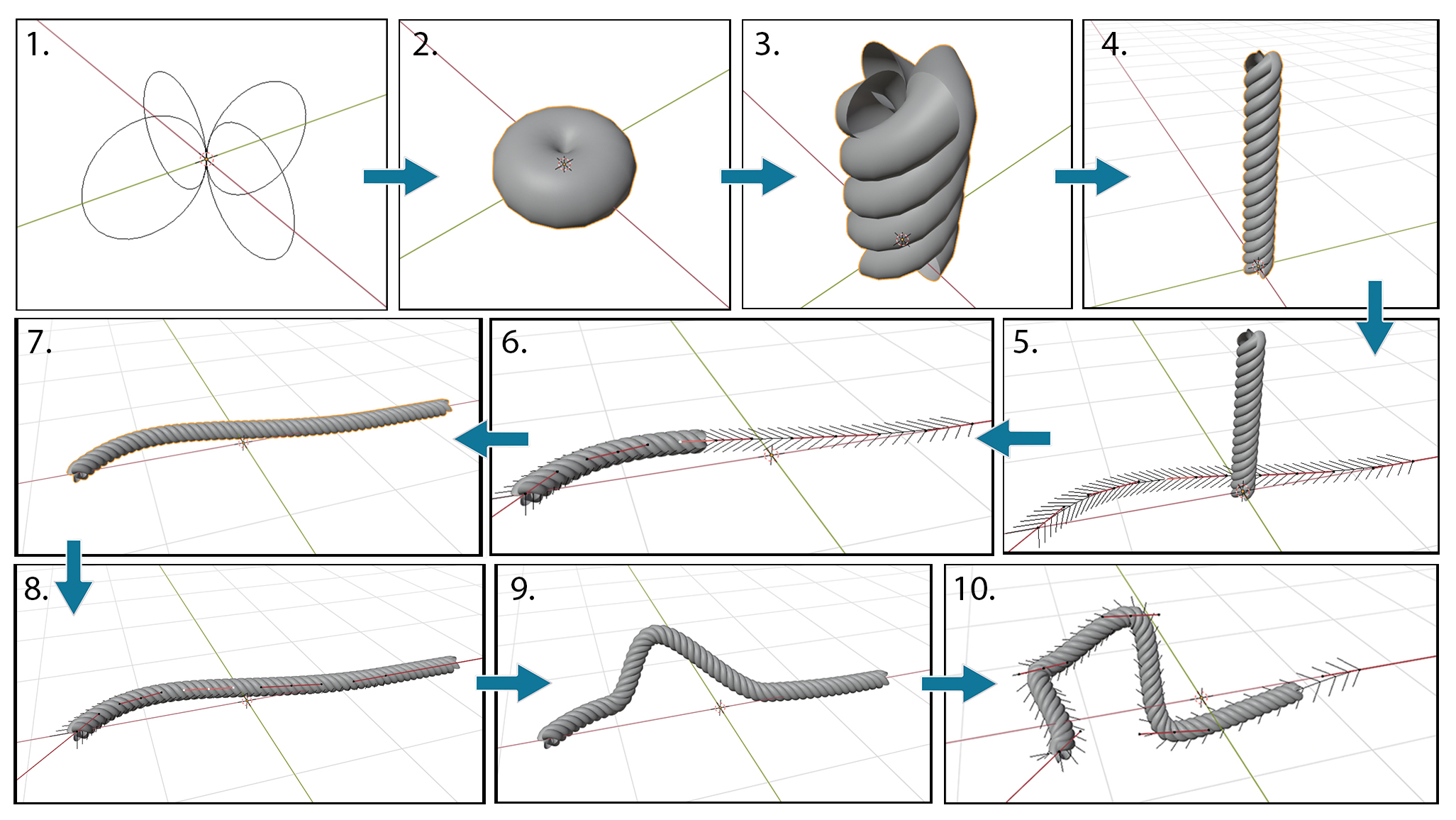}
\caption{Blender rope modelling pipeline, described in \cite{blendertutorial}. 1: Four circles are joined as one mesh. 2: The 'Screw Modifier' is applied to the mesh to produce a helix-like braided appearance. 3-4: the rope is elongated by increasing the 'Screw Length' and 'Iterations' attributes. 5: A Bezier curve is added to the scene. The black points represent control points and the orange segments are handles for the control points. The handles and control points structure the shape of the Bezier curve. 6: The Bezier curve is added as a 'Curve Modifier' to the rope mesh along the Z-axis, so that the mesh will traverse along the curve. 7: The rope is further elongated. 8-9-10: Control points along the Bezier curve can be freely displaced in 3D space to deform the rope, or can be arranged to produce a specific configuration.}
  \label{rope-design-fig}
\end{figure}

The rope is modeled as a deformable four-stranded braid as in Figure \ref{rope-design-fig}, resembling twisted nylon ropes commonly used in lifting/towing/pulley applications. A sphere is added on one side to disambiguate both ends of the rope.

\subsubsection{Deformation}
\begin{figure}[!htbp]
  \includegraphics[width=1.0\linewidth]{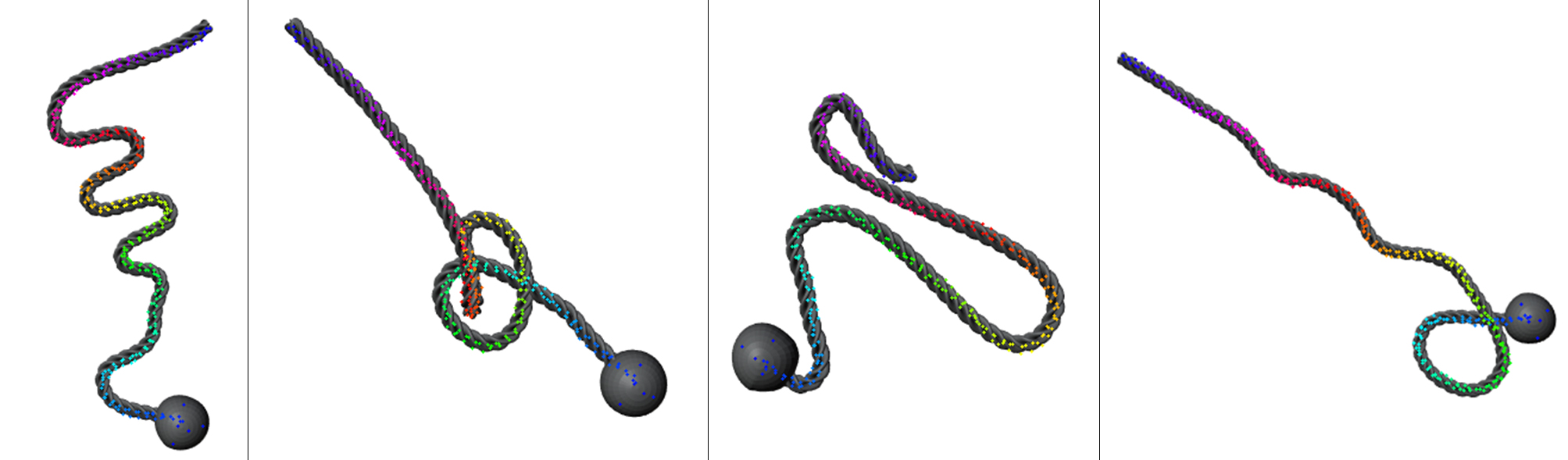}
\caption{Ground truth annotations. Using the transform between Blender's world coordinate system and the simulated camera used in rendering, we collect dense, semantically consistent pixel-wise annotations of the rope in various configurations. }
  \label{loop-knot-fig}
\end{figure}

Arbitrary planar deformations are generated by randomly displacing the x and y coordinates of a subset of the rope's Bezier control points, given by ${\textbf{P}_0, ..., \textbf{P}_\textnormal{M-1}}$. We simulate more complex nonplanar rope configurations by geometrically arranging ${\textbf{P}_0, ..., \textbf{P}_\textnormal{M-1}}$ to yield the desired curvature in the shape of loops and knots as shown in Figure \ref{loop-knot-fig}.

\begin{figure}[t!]
  \includegraphics[width=1.0\linewidth]{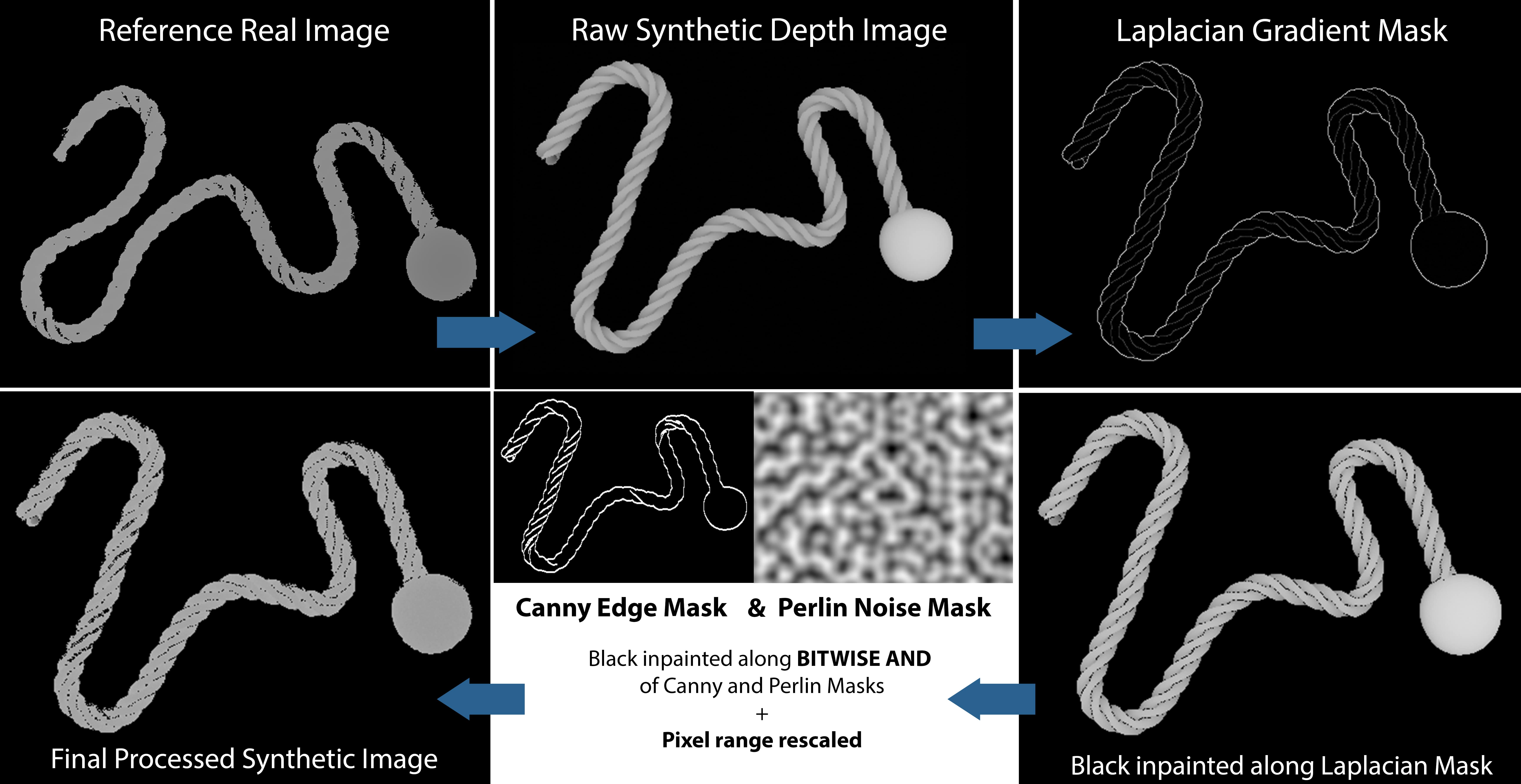}
  \caption{Sim-to-Real Processing Pipeline. A raw synthetic depth image is post-processed to look like a real reference depth image (top left) by scaling the pixel range and strategically inpainting black along noise, edge, and gradient masks as described in \ref{sec:sim2real}. This post-processing of the simulation images models the noise and black pixel corruption in real depth images along regions of high gradient. A DDOD mapping is trained on these processed simulated images to enable sim-to-real transfer.}
  \label{sim2real-fig}
\end{figure}

\subsubsection{Domain Randomization}
\label{sec:sim2real}
We leverage several domain randomization and image processing techniques to enable sim-to-real transfer by training on rendered synthetic depth images that are post-processed to match real images. In simulation, we slightly randomize over rope hyperparameters in the specified ranges from Table \ref{tab:blender-params} (all in Blender standard units). The intent is to account for slight dimension mismatch between domains. Additionally, we inject both zero-mean, unit variance Gaussian and Poisson noise in the simulated images to model the noise in real depth images. In real depth images of the rope, the corrupted pixels tended to occur along regions of high gradient, particularly on braided rope contours. To model this in the simulated images, we randomly color pixels black along areas of high Laplacian gradient and edges detected with a Canny edge detector \cite{bao2005canny} on the rope along a Perlin noise mask \cite{perlin2002improving}. The Canny edges and Laplacian gradient provide rope contours and the Perlin noise provides realistic gradient noise. We rescale the pixel range of the simulated images to match the pixel range in real given a single reference real depth image. This process is illustrated in Figure \ref{sim2real-fig}.

\subsection{Gradient Tracking Details}
We describe the implementation of an analytical method for acquiring rope correspondences from Section \ref{sec:results}. This method aims to trace along the length of the rope from one endpoint to the other, recording ordered pixel annotations along the way. Examining the \emph{pixel intensity gradient} for the local patch at each step in the tracing process yields the direction to step next.
Given a segmentation mask of the rope in an arbitrary planar configuration, we use the gradient-based method to find correspondences as an analytical alternative to a descriptor-based approach. The full method can be broken down into three steps: pre-processing the images, acquiring annotations by tracing orthogonally to the gradient along the rope, and doing matching between a pair of images given a set of ordered annotations for each image.

\subsubsection{Image Preprocessing Pipeline}
We apply OpenCV inpainting to the input segmentation mask of the rope to fill missing pixels using nearby pixels in the vicinity. Next, we use Gaussian blurring followed by binary thresholding to smooth the edges of the rope. This step is meant to ensure that small frays in the rope do not affect the overall gradient within a crop. To find the starting point on the rope, we use OpenCV-based Hough circle detection to locate the center of the attached ball. Given this point as a reference, the starting point of the rope is considered to be the nearest-neighbor on the rope mask, outside a fixed radius from the ball center. 

\algnewcommand{\LineComment}[1]{\State \(\triangleright\) #1}
\MakeRobust{\Call}
\begin{algorithm}[H]
\label{alg:preprocess_alg}
\caption{Image Preprocessing Pipeline}\label{alg:preprocess}
\begin{algorithmic}[1]

\Procedure{preprocess}{img, circle\_radius}
\State img $\gets$ \Call{inpaint}{img}
\State img $\gets$ \Call{gaussian\_blur}{img}
\State img $\gets$ \Call{binary\_threshold}{img}
\State circle\_center $\gets$ \Call{hough\_circles}{img}
\LineComment{\emph{Find starting point on the rope circle\_radius + 5 pixels away from the circle center}}
\State angle $\gets$ 0
\While{TRUE}
\State dx $\gets$ (circle\_radius + 5) * \Call{sin}{angle}
\State dy $\gets$ (circle\_radius + 5) * \Call{cos}{angle}
\State rope\_start $\gets$ circle\_center + [dx, dy]
\If{img[rope\_start] > 0} 
\State \textbf {break} 
\EndIf
\State angle $\gets$ angle + 5
\EndWhile
\State \textbf{return} img, rope\_start, circle\_center
\EndProcedure
\end{algorithmic}
\end{algorithm}

\subsubsection{Gradient Annotation Algorithm}
The algorithm for accumulating annotations by following the pixel intensity gradient is presented in Algorithm \ref{alg:grad}.
\algnewcommand\algorithmicswitch{\textbf{switch}}
\algnewcommand\algorithmiccase{\textbf{case}}
\algdef{SE}[SWITCH]{Switch}{EndSwitch}[1]{\algorithmicswitch\ #1\ \algorithmicdo}{\algorithmicend\ \algorithmicswitch}%
\algdef{SE}[CASE]{Case}{EndCase}[1]{\algorithmiccase\ #1}{\algorithmicend\ \algorithmiccase}%
\algtext*{EndSwitch}%
\algtext*{EndCase}%

\MakeRobust{\Call}
\begin{algorithm}[H]
\label{alg:gradient_alg}
\caption{Gradient Annotation Algorithm}\label{alg:grad}
\begin{algorithmic}[1]

\Procedure{gradient}{crop}
\State grad\_x, grad\_y $\gets$ \Call{np.gradient}{crop}
\State directions $\gets$ [0, 0, 0, 0]
\LineComment{\emph{Find which direction (up, down, right, left) most of the pixel intensity gradients point}}
\For{$i \in \{0,\dots,len(grad\_x)\}$}
  \For{$j \in \{0,\dots,len(grad\_x[0])\}$}
    \State temp\_grad $\gets$ [grad\_x[i][j], grad\_y[i][j]]
    \State directions $\gets$ \Call{update}{directions, temp\_grad}
    \LineComment{\emph{Update directions with direction that temp\_grad points}}
  \EndFor
\EndFor
\If{directions[2] > directions[3]}
    \State final\_x $\gets$ directions[2]
\Else
    \State final\_x $\gets$ -1 * directions[3]
\EndIf
\If{directions[0] > directions[1]}
    \State final\_y $\gets$ directions[0]
\Else
    \State final\_y $\gets$ -1 * directions[1]
\EndIf
\State \textbf{return} \Call{normalize}{[final\_x, final\_y]}
\EndProcedure

\Procedure{update\_step}{direction, curr\_x, curr\_y, step\_size}
\State dir\_x, dir\_y $\gets$ direction
\State next\_x $\gets$ curr\_x + dir\_x $\times$ step\_size
\State next\_y $\gets$ curr\_y + dir\_y $\times$ step\_size
\State \textbf{return} next\_x, next\_y
\EndProcedure

\Procedure{follow\_grad}{img, step\_size, crop\_size}
\State img, rope\_start, circle\_center $\gets$ \Call{preprocess}{img}
\State curr\_x, curr\_y $\gets$ rope\_start
\State steps\_taken $\gets$ 0
\State points $\gets$ [  ]

\LineComment{\emph{Follow gradient orthogonally until max number of steps reached (empirically determined, approximates end of rope)}}
\While{steps\_taken \textless MAX\_NUM\_STEPS}
\State points.append([curr\_x, curr\_y])
\State crop $\gets$ \Call{local\_crop}{img, curr\_x, curr\_y, crop\_size}
\LineComment{\emph{Take a 2 * crop\_size by 2 * crop\_size crop of the image centered at curr\_x and curr\_y}}
\State grad\_x, grad\_y $\gets$ \Call{gradient}{crop}
\State direction $\gets$ \Call{adjust}{\Call{orthogonal}{grad}}
\LineComment{\emph{ADJUST flips direction if it is towards, instead of away, from (curr\_x, curr\_y)}}
\State curr\_x, curr\_y $\gets$ \Call{update\_step}{direction, curr\_x, curr\_y, step\_size}
\State steps\_taken $\gets$ steps\_taken + 1
\EndWhile\label{gradientendwhile}
\State \textbf{return} points\Comment{The final annotations list}
\EndProcedure
\end{algorithmic}
\end{algorithm}

\subsubsection{Matching}
To acquire $N$ correspondences between a pair of images, we first compute their gradient-based annotation lists independently using the gradient annotation algorithm. Then, we subsample every $\lceil\frac{L}{N}\rceil^{\textnormal{th}}$ pixel, where $L$ is the length of the annotation list. 

\begin{figure*}[!htbp]
\centering
  \includegraphics[width=0.8\linewidth]{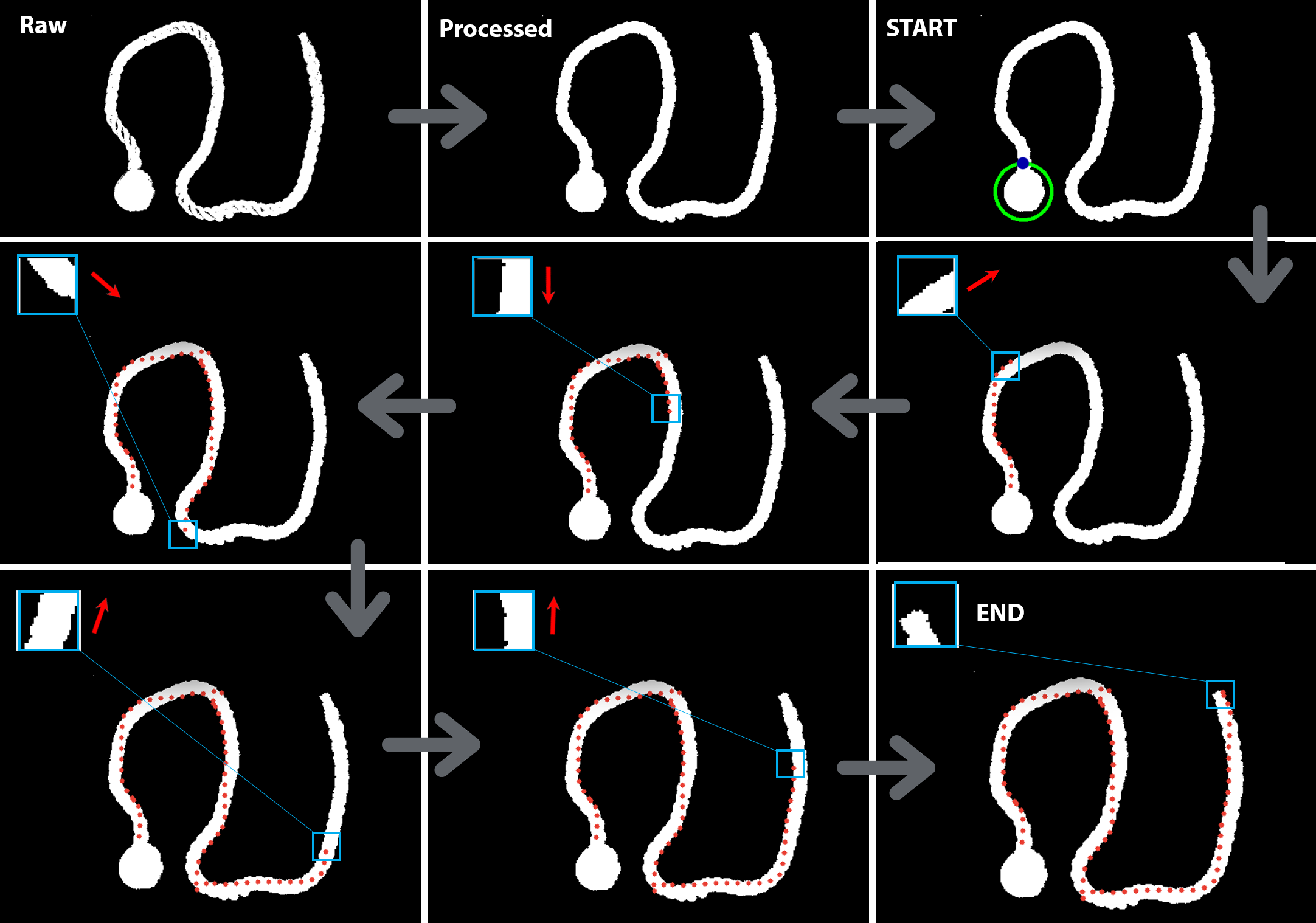}
  \caption{Gradient-based baseline correspondence method. Given a raw segmentation mask of the rope, we preprocess it by inpainting the black artifacts and low-pass filtering using OpenCV functions. Next, we locate a starting annotation (blue). To do this, we use OpenCV's hough circles to find the center of the ball, and use this point as a reference to find the start of the rope. We choose this point because the circle is a consistent and localizable feature to find across images using classical methods. From the start point, we take a local crop, compute the pixel intensity gradient of the crop, and take a step in the direction orthogonal to the gradient. We re-sample the crop and repeat for a fixed number of steps until the end of the rope is reached.}
  \label{grad-fig}
\end{figure*}

\subsubsection{Failure Cases}
\begin{figure}[!htbp]
  \includegraphics[width=1.0\linewidth]{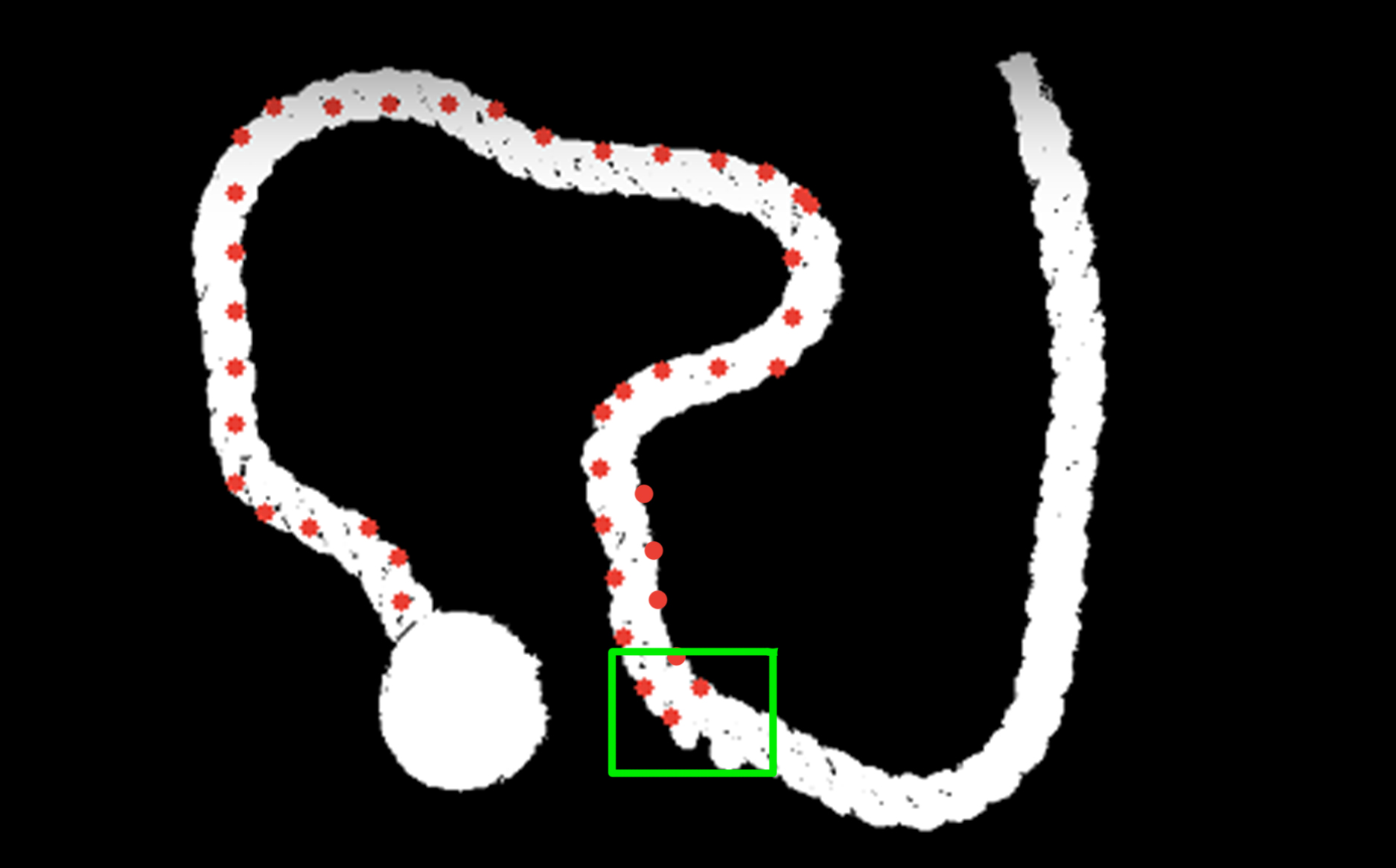}
\caption{Rope irregularities failure mode. The nylon rope has non-uniform thickness and fraying. This can confuse the gradient direction, causing the annotations to follow along the curve of an irregular bump rather than along the rope geometry.}
  \label{bump-fail-fig}
\end{figure}

Irregularities in rope thickness that persist after image preprocessing may result in a pixel intensity gradient that is not orthogonal to the direction of the overall rope (Figure \ref{bump-fail-fig}). When this occurs, the algorithm takes steps along the width of the rope rather than the length until the opposite side of the rope is reached. Then, the algorithm will continue taking steps along the length of the rope but in the opposite direction, resulting in a loop backwards where the annotated points are now overlapping with previous annotations. 

\begin{figure}[!htbp]
  \includegraphics[width=1.0\linewidth]{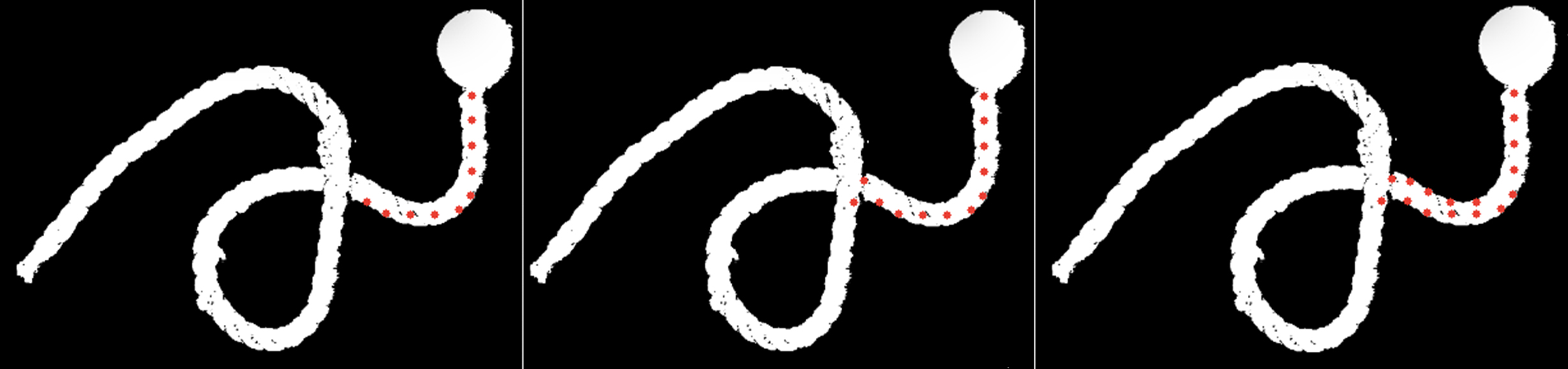}
\caption{Nonplanar failure mode. One failure of the analytical-based approach is that the  gradient is unclear at areas of overlap in the rope. In this case, the gradient does not follow the rope geometry but rather starts collecting annotations in the opposite direction.}
  \label{nonplanar-fail-fig}
\end{figure}

The gradient annotation algorithm is currently unstable for nonplanar configurations of the rope. With crops containing self-occluding regions of the rope, it is unclear where the pixel intensity gradient will point, leading to confusion in annotations at the point of overlap (Figure \ref{nonplanar-fail-fig}). In future work, we will consider using the depth image gradient (as opposed to the segmentation mask gradient) which may provide richer information to support these cases.

\subsection{Experimental Details}
\subsubsection{Physical Experiment Setup}
The pipeline of taking raw depth observations provided by the Photoneo Phoxi 3D Scanner and doing manipulation from these observations is as follows:
\begin{itemize}
    \item We acquire a raw point cloud of the rope on top of the surface used in manipulation.
    \item Using pre-computed workspace boundaries in the world frame, we pre-process the resulting point cloud by removing points that lie on the manipulation surface or above the rope. This is to ensure that the final image of the workspace has a clean segmentation of the rope, since we did not domain-randomize the background of images seen in training.
    \item We project the point cloud from the world frame to camera frame using a known world-to-camera calibration acquired via a chessboard. This results in a grayscale image of size 772 $\times$ 1032.
    \item We downscale this image to size 640 $\times$ 480 (the dimensions used for images in training) before passing the image through the neural network, and get pixel-wise correspondences via matching in descriptor space.
    \item We upscale all pixel annotations to match the 772 $\times$ 1032 scale of the original image.
    \item Finally, the robot end-effector can grasp at a point in world space given the properly scaled pixel annotations using the camera-to-world transformation.
\end{itemize}
\subsubsection{One-Shot Visual Imitation Details}
We elaborate on the experimental setup for the one-shot visual imitation policy described in Section \ref{sec:imitation-policy}.
\begin{itemize}
    \item A human demonstrator records a sequence of images separated by one pick-and-place action to deform the rope.
    \item In order to "imitate" the sequence, the robot uses the \emph{last} recorded image, which captures the workspace after the human demonstration, and sequentially uses the previously recorded image as the next subgoal in the demonstration until the first recorded image is reached. 
    \item In this fashion, the robot repeatedly "undoes" the set of actions the human demonstrator did, doing reverse imitation, rather than imitation in sequence. This is because imitation in sequence would require that the human and robot start from the exact same rope configuration, which is not possible after the human demonstrator has already deformed the rope during the demonstration.
\end{itemize}

\subsection{Hyperparameters}
\begin{table}[!htbp]
    \caption{Rope Simulator Hyperparameters.}
    \centering
    \begin{tabular}{||c || c ||} 
 \hline
 Parameter & Range of Values\\ 
 \hline\hline
rope thickness & [0.05, 0.065] \\ 
 \hline
rope length & [14.3, 15] \\ 
 \hline
coil length (length of braid texture) & [12.5, 14]\\ 
 \hline
attached sphere radius & [0.35, 0.37] \\ 
 \hline
\end{tabular}

\label{tab:blender-params}
\end{table}

\begin{table}[!htbp]
    \caption{Descriptor Training Hyperparameters.}
    \centering
    \begin{tabular}{||c || c ||} 
 \hline
 Parameter & Value(s)\\ 
 \hline\hline
 number of training images & 3600 \\ 
 \hline
 learning rate & 1.0$\textnormal{e}^{-4}$ \\ 
 \hline
 learning rate decay & 0.9 \\
 \hline
 steps between learning rate decay & 250 \\
 \hline
 training iterations & 3500 \\
 \hline
 descriptor contrastive margin ($M$) & 0.5 \\
 \hline
 descriptor dimension & (3,6,9,16) \\
  \hline
 number of annotations & 500-1600 \\
  \hline
\end{tabular}

\label{tab:training-params}
\end{table}

\begin{table}[!htbp]
    \caption{One-Shot Visual Imitation Hyperparameters.}
    \centering
    \begin{tabular}{||m{0.3\linewidth} || m{0.4\linewidth} || m{0.1\linewidth} ||} 
 \hline
 Parameter & Explanation & Value\\ 
 \hline\hline
inter-annotation distance & distance between sparsely sampled pixels on rope mask & 50px \\ 
 \hline
number of nearest neighbors & number of matches sampled per annotation (median of these is taken to be final correspondence) & 100 \\ 
 \hline
IoU thresh & intersection-over-union threshold between current workspace image and goal image that determines when to move to next subgoal & 0.67 \\
 \hline
subgoals & number of steps in the demonstration & 4 \\
 \hline
attempts per subgoal & number of robot actions to reach each subgoal & 3 \\
 \hline
\end{tabular}

\label{tab:manip-params}
\end{table}

\begin{figure*}[!htbp]
\centering
  \includegraphics[width=1.0\linewidth]{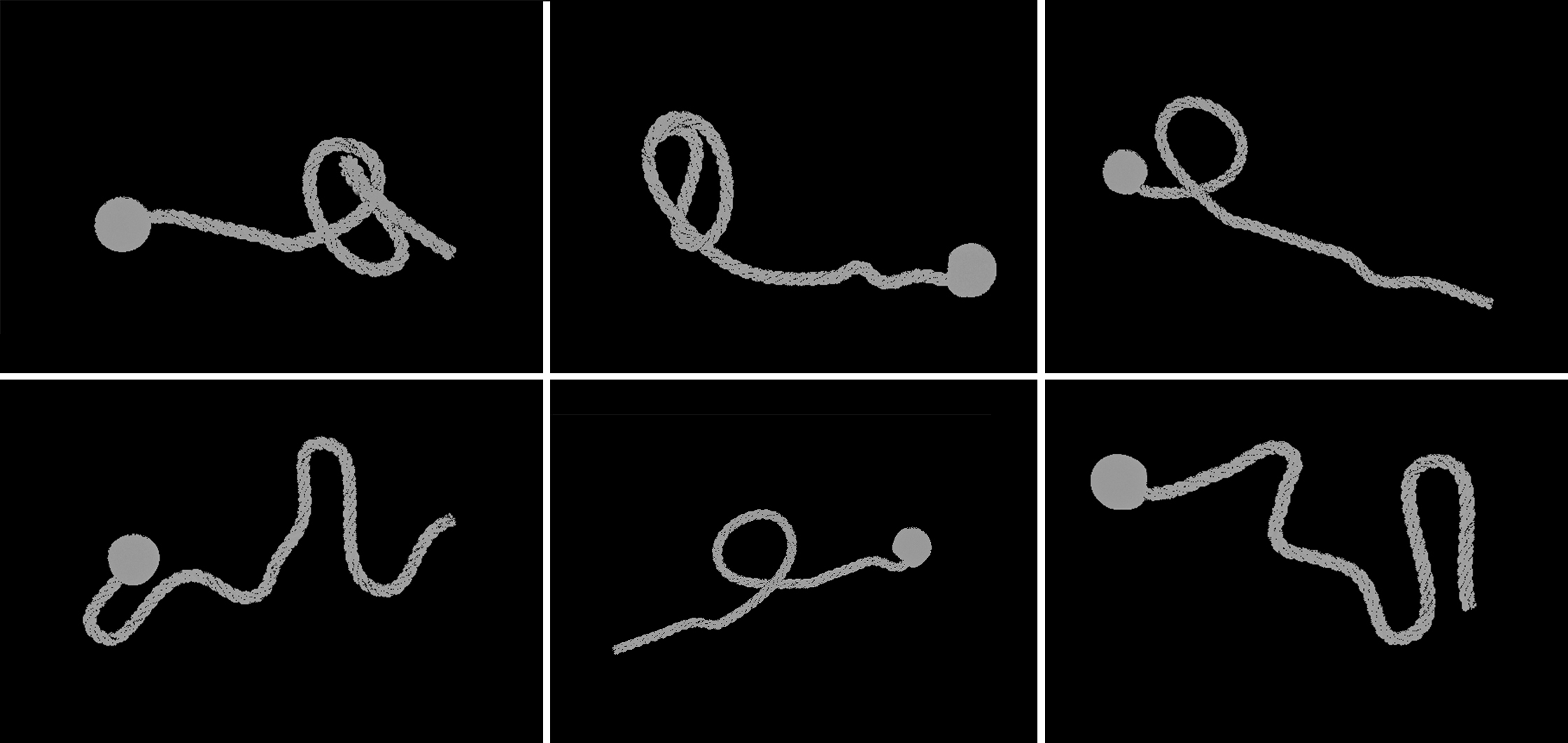}
  \caption{Representative examples of both planar and non-planar simulated depth images rendered with Blender, with noise injection to model missing pixels in real images.}
  \label{sim-example-fig}
\end{figure*}

\begin{figure*}[!htbp]
\centering
  \includegraphics[width=1.0\linewidth]{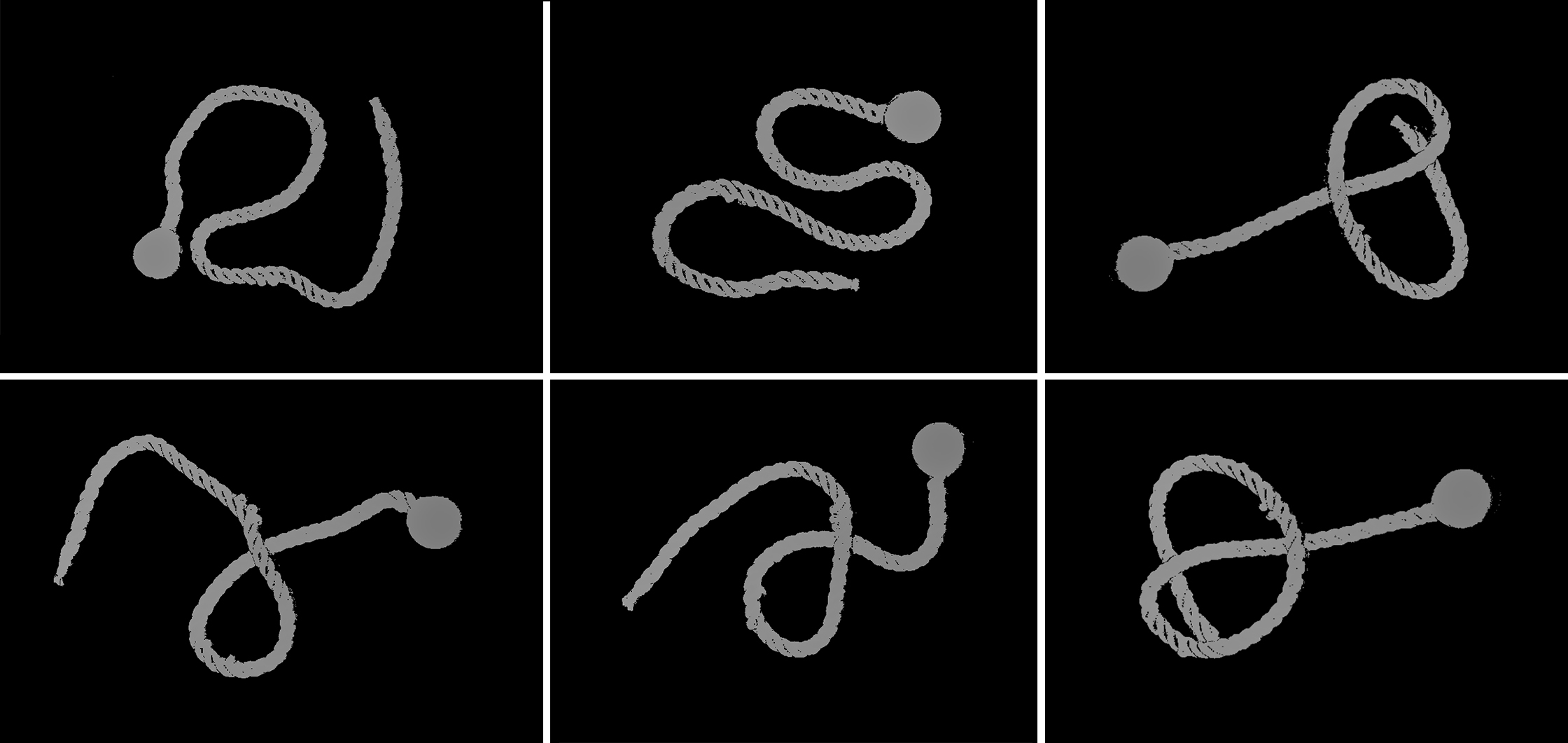}
  \caption{Representative examples of both planar and non-planar real depth images.}
  \label{real-example-fig}
\end{figure*}

\begin{figure*}[!htbp]
  \includegraphics[width=0.95\textwidth,height=0.95\textheight,keepaspectratio]{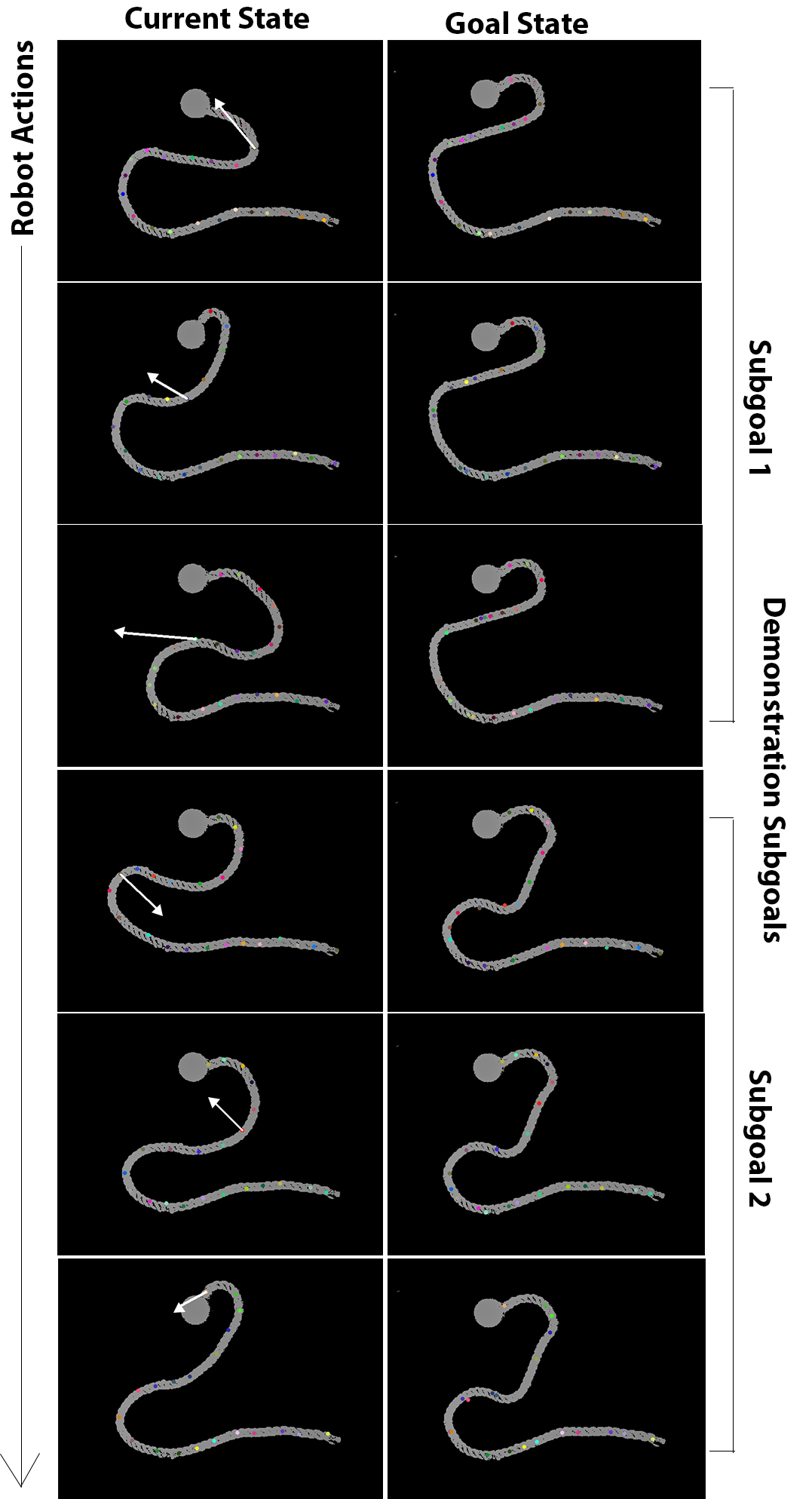}%
  \caption{Representative example of six actions in one rollout of the one-shot visual imitation policy (Section \ref{sec:visual-imitation-exps}). The robot attempts to achieve two subgoals in a given demonstration, with three pick-and-place attempts per subgoal. Left column (from top to bottom): actions planned and taken by robot. Right column: two subgoals from human demonstration.}
  \label{imitation-rollout-fig}
\end{figure*}

\end{document}